\documentclass[letterpaper, 10 pt, conference]{ieeeconf}  

\IEEEoverridecommandlockouts                              

\usepackage{graphics} 
\usepackage{epsfig} 
\usepackage{amsmath} 
\usepackage{amsfonts}
\usepackage{amssymb}  
\usepackage{mathtools, mathdots}
\usepackage{algorithmic,algorithm}
\usepackage{subcaption}
\usepackage{multirow}
\usepackage{booktabs} 
\usepackage{textcomp}
\usepackage{svg}
\usepackage[labelfont=bf,
            belowskip=0\baselineskip, 
            aboveskip=0\baselineskip,
            skip=0\baselineskip]{caption}
\usepackage{hyperref}
\hypersetup{
    colorlinks=true,
    linkcolor=blue,
    filecolor=magenta,
    urlcolor=blue,
}

\definecolor{objects}{rgb}{0.09, 0.75, 0.82}
\definecolor{table}{rgb}{1.0, 0.5, 0.01}
\definecolor{cabinet}{rgb}{0.122, 0.467, 0.706}
\definecolor{wall}{rgb}{0.682, 0.780, 0.902}
\definecolor{clothes}{rgb}{0.807, 0.427, 0.745}
\definecolor{mirror}{rgb}{0.859, 0.859, 0.5449}
\definecolor{picture}{rgb}{0.839, 0.153, 0.157}
\definecolor{lighting}{rgb}{0.710, 0.812, 0.416}

\definecolor{ceiling}{rgb}{0.612, 0.613, 0.875}
\definecolor{appliances}{rgb}{0.906, 0.796, 0.588}
\definecolor{sink}{rgb}{0.518, 0.235, 0.224}
\definecolor{stool}{rgb}{0.388, 0.475, 0.243}
\definecolor{plant}{rgb}{0.549, 0.635, 0.314}
\definecolor{counter}{rgb}{0.647, 0.318, 0.588}
\title{\LARGE \bf
PEACE: Prompt Engineering Automation for CLIPSeg Enhancement for Safe-Landing Zone Segmentation
}

\author{Haechan Mark Bong$^{1, 2}$,
        Rongge Zhang$^{1}$,
        Antoine Robillard$^{1}$,
        Giovanni Beltrame$^{1, 2}$
        \thanks{$^{1}$Department of Computer Engineering and Software Engineering, Polytechnique Montréal, QC., Canada. {\tt\small haechan.bong@polymtl.ca}}
        \thanks{$^{2}$MILA, QC., Canada.}
        \thanks{This work was supported by the National Research Council Canada.}
}

\begin{document}

\maketitle
\thispagestyle{empty}
\pagestyle{empty}

\begin{abstract}
Safe landing is essential in robotics applications, from industrial settings to space exploration. As artificial intelligence advances, we have developed PEACE (Prompt Engineering Automation for CLIPSeg Enhancement), a system that automatically generates and refines prompts for identifying landing zones in changing environments.
Traditional approaches using fixed prompts for open-vocabulary models struggle with environmental changes and can lead to dangerous outcomes when conditions are not represented in the predefined prompts. PEACE addresses this limitation by dynamically adapting to shifting data distributions. Our key innovation is the dual segmentation of safe and unsafe landing zones, allowing the system to refine the results by removing unsafe areas from potential landing sites. Using only monocular cameras and image segmentation, PEACE can safely guide descent operations from 100 meters to altitudes as low as 20 meters. The testing shows that PEACE significantly outperforms the standard CLIP and CLIPSeg prompting methods, improving the successful identification of safe landing zones from 57\% to 92\%. We have also demonstrated enhanced performance when replacing CLIPSeg with FastSAM. The complete source code is available as an open-source software~\footnote{\hyperlink{https://github.com/MISTLab/PEACE}{https://github.com/MISTLab/PEACE}}.
\end{abstract}

\section{Introduction}
Logistics is a pivotal element in diverse sectors, from e-Commerce operations to complex military operations. The use of unmanned aerial vehicles (UAVs)~\cite{garber2012style,koren_2019} in logistics is gaining both research and industrial attention. In this landscape, autonomous robots have emerged as a widely sought-after solution. Particularly in modern urban environments, aerial robots offer a promising approach to improve last-mile delivery efficiency and reduce carbon emissions. However, safety concerns have significantly limited the widespread adoption of flying robots in densely populated areas. When not properly designed and operated, these devices can pose potential threats to structures, vehicles, and the general public, especially if issues with geolocation or other sensory information prevent safe landing. Our aim is to achieve secure emergency landings without the need for external communication or infrastructure, using only the computational capabilities and perceptual abilities of compact and lightweight cameras onboard.

The primary objective of safe self-landing for UAVs involves identifying and descending into a designated safe-landing zone (SLZ), such as relatively flat grasslands or open fields, while avoiding pedestrians, vehicles, and structures. This capability becomes especially critical during localization (e.g., GPS) or remote control communication failures, where a successful landing ensures that operators can eventually regain control of the device. Current automatic landing systems typically rely on traditional localization and perception methods using Simultaneous Localization and Mapping (SLAM)—which are limited by sensor performance and computational resources—or conventional deep learning-based image segmentation models that struggle with domain adaptation challenges.

Our previous work, DOVESEI~\cite{bong2023dynamic}\footnote{Only available as an extended abstract.}, introduced a system capable of operating with only a monocular RGB camera that can "dynamically focus" by masking raw segmentation according to the current state of the system. DOVESEI leverages advances in large language models~(LLMs) to utilize Open-Vocabulary (OV) models, allowing for language-based "tuning" without extensive data collection. While LLMs for aerial robotics remains a relatively underdeveloped research area, DOVESEI's innovation was incorporating an OV model for segmentation with dynamic focusing to enhance overall system performance. The system is built on CLIPSeg (CS)~\cite{Luddecke_2022_CVPR}, which requires appropriate prompts for optimal performance, making SLZ selection quality directly dependent on prompt engineering. Drawing from CLIP~\cite{radford2021learning}, which demonstrated improved results using "A photo of {}" prompts (where {} represents target-defining terms), DOVESEI used CLIP Interrogator~\cite{pharmapsychotic2022} to heuristically generate better prompts for aerial imagery. However, this aerial prompt engineering remained static, typically created through trial and error by selecting terms that describe aerial environments, and applied uniformly across all input images. For safe-landing applications, this static approach is inadequate when faced with constantly changing aerial perspectives that introduce data distribution shifts.

Using a single prompt generated by CLIP Interrogator does not effectively represent the dynamic nature of aerial imagery. To address this limitation, we propose PEACE (Prompt Engineering Automation for CLIPSeg Enhancement), designed to enhance model adaptability and stability in evolving real-world scenarios. PEACE dynamically generates optimized prompts specific to each input image, representing a significant advancement toward more robust autonomous UAV systems. Our main contributions through PEACE include two key improvements:

\begin{enumerate}
    \item Dynamic aerial prompt engineering per image frame that can adapt to changing environments during safe-landing zone segmentation.
    \item A novel method for improving safe-landing zone segmentation accuracy by combining positive and negative term segmentations: First, we generate segmentations for safe-landing zones using positive terms (grass, park, water, etc.) and merge them together. Similarly, we generate and merge segmentations for unsafe-landing zones using negative terms (concrete, street, building, etc.). Finally, we refine landing zone segmentation by removing areas identified as unsafe, resulting in more accurate and reliable landing zone identification.
\end{enumerate} 

We target aerial segmentation at an altitude between 100 and 20 meters, the expected altitude of UAV logistics corridors. Most of the previous work operates at altitudes of up to 30 meters~\cite{9981152,chatzikalymnios2022,7138988,Mittal2018VisionbasedAL,6884813} to align with the capabilities of small stereo cameras. Similarly to DOVESEI~\cite{bong2023dynamic}, our work offloads the remaining 20m to be navigated using conventional 3D path planning methods, since the objective of our work is to improve aerial segmentation in dynamic environments, not in path planning. Experiments on testing PEACE in DOVESEI setting presented segmentation enhancements which took advantage of dynamically adapting prompts to images compared to other methods.

\section{Related Work}

\subsection{Safe Self-landing for UAV}
Many previous systems dealt with automatic UAV landing, but they would limit the maximum altitude~\cite{chatzikalymnios2022,7138988,Mittal2018VisionbasedAL,6884813,rabah2018} (due to the small baseline, of 3D cameras, limiting their functional maximum depth range), use more expensive and heavier sensors~\cite{cesetti2010,6836176,rabah2018} (e.g. 3D LiDARs), or scan and create a map of the environment using Simultaneous Localization and Mapping (SLAM)~\cite{9981152,chatzikalymnios2022,Mittal2018VisionbasedAL} before deciding where to land. Our method relies on a map of the environment or odometry estimation and depends only on the images captured by the camera. In addition, an inherent advantage of the proposed methodology lies in its adaptability to diverse scenarios by not targeting a specific safe-landing zone such as flat surface, road or grass. Requiring minimal parameter adjustments, our approach can cater to varying environments and operational conditions without necessitating extensive data collection or recalibration.

\subsection{Aerial Image Segmentation}
Past use of segmentation models~\cite{long2015fully,ronneberger2015u} has demonstrated a significant limitation: these systems struggle to tolerate domain shifts, as they were trained and tested only for very specific scenarios. This limitation has fueled growing interest in Vision-Language models (VLMs), which exhibit promising zero-shot performance, enabling them to generalize across scenarios and adapt to domain shifts. The key factor for the success of these models lies in accurate prompts, which require meticulous, usually non-obvious, textual adjustments and proper design. In this process, prompt engineering is crucial, as we observe that even a minor variation in wording can ultimately have a noticeable positive or negative impact on final segmentation performance. Inspired by prompt learning in language tasks, Du et al.~\cite{du2022learning} introduced context optimization to automate prompt engineering for small-sample classification. Recent work~\cite{zhou2022learning} extended~\cite{du2022learning} to Open-Vocabulary (OV) object detection, devising fine-grained automatic prompt learning and specialized background explanations to identify the necessary prompts. OV object detection uses natural language to detect any objects, enabling domain shifts. However, these methods~\cite{du2022learning,zhou2022learning} are primarily tailored for specific object detection tasks and yield sub-optimal results when applied to aerial image segmentation. Currently, research efforts are focused in generalized OV segmentation and~\cite{kirillov2023segment, zhao2023fast} demonstrated promising results when segmenting any areas within an image, but they are yet to be tested on low resolution and images taken at an high altitude.   

\subsection{Open-Vocabulary Model for Landing Zone Selection}
\begin{figure*}[htbp]
\centerline{\includesvg[inkscapelatex=false, width=1\linewidth]{img/PEACE_DP.svg}}
\caption{PEACE Data Pipeline: An observed aerial image is passed through a modified CLIP Interrogator to generate a best-matching word for each categories in the description type (natural environment, image resolution and image type (frames)). The selected words are inserted into the optimized prompt (\emph{"A \{Res.\} \{Frame\} of \{pos. or neg. term\} in \{env.\}"}) to complete the prompt. This prompt and the image are used as inputs to an open-vocabulary segmentation model and generates a segmentation targeting positive (pos.) or negative (neg.) term.}
\label{methods:PEACE_pipe}
\end{figure*}
In our previous work~\cite{bong2023dynamic}, we introduced an OV model, DOVESEI, which demonstrates safe-landing capabilities through OV segmentation and aerial prompt engineering without requiring extensive data collection. Using CLIPSeg~\cite{Luddecke_2022_CVPR}, DOVESEI achieved adaptability to various safe-landing target classes with minimal parameter adjustments, eliminating the need for extensive data accumulation to refine internal models. While numerous other OV models exist, we selected CLIPSeg because it is not only one of the early OV models that is simple to reproduce, but also lightweight, which is a critical requirement for UAV deployment. However, a significant limitation of DOVESEI is its reliance on heuristically selected static prompts to describe all images regardless of their content. Our hypothesis is that optimal SLZ segmentation results and appropriate prompt selection are highly interdependent, and superior performance can only be achieved through dynamic prompts that adapt to changing environments and image content.

For our current work, we also experimented with other popular and computationally efficient OV segmentation models, such as FastSAM (FS)~\cite{zhao2023fast}, to compare the results with CLIPSeg and demonstrate that PEACE can adapt to any OV segmentation model and improve the accuracy of the segmentation for aerial images. We selected FastSAM not only because it is a widely-used model that facilitates reproducibility but also because of its speed, which is essential for on-board systems mounted on UAVs.

\section{System Design}
\subsection{PEACE Prompt Generation}
The selection of the best landing locations (in pixels) is performed using Open-Vocabulary (OV) segmentation models. Our previous work DOVESEI used aerial prompt engineering: \emph{"A bird's eye view of a \{term\}, in game screenshot, bad graphics, shade, shadows."} with CLIPSeg, where \emph{\{term\}} is the target class to segment.  \emph{\{term\}} can have varying definitions, but our definition of a safe-landing zone was an area that is easily accessible, far from human and being destroyed by cars or trees. This well-selected combination of words that describe aerial images was chosen heuristically after trial and error to align with the general environment of aerial view at high altitude compared to the baseline (default) prompt \emph{"A photo of \{term\}"} used by CLIPSeg. However, this prompt is static and not optimal since every image per frame changes during flight, which results in changes in the environment observed by CLIPSeg. 

To address this problem, we proposed a PEACE data pipeline, shown in Fig.~\ref{methods:PEACE_pipe}. In this pipeline, we modified the CLIP Interrogator~\cite{pharmapsychotic2022} to specifically generate automated aerial prompt engineering. CLIP Interrogator is designed to generate captions based on an image, which describes everything observed within an image in an artistic fashion. In addition, it uses BLIP~\cite{li2022blip} to generate detailed captions to maximize the precision of the image description. The caption is generated using best-matching words by ranking words based on word-to-image similarity (CLIP's cosine similarity~\cite{radford2021learning}) from description types which are categories such as artists, mediums, movements, trending and flavors and each category consists of words that are used to describe the corresponding category. For PEACE, we replaced these lists of description types with natural environments (e.g. sunny, rainy, winter, foggy, etc.), image resolution (e.g. good, bad, blurry, etc.) and image type (e.g. a screen, photo, image, view, etc.). We experimented with description types such as aerial (high, top, altitude, etc.) and context-aware (next to, top of, behind, etc.), but generally, they were not improving SLZ segmentation. BLIP captions inside CLIP Interrogator were removed from prompts since our objective is to target a selection of potential SLZs and BLIP's description of every single detail in an image was excessive and not helpful for segmentation. After experimenting with prompt engineering in a heuristic manner, we realized that optimal segmentation was obtained from the prompt \emph{"A \{resolution\} \{image type (frame)\} of \{term\} in \{environment\}}." which is the prompt used in PEACE and \emph{\{resolution\}, \{image type (frame)\}} and \emph{\{environment\}} are selected words from our modified CLIP Interrogator based on an input image. \emph{\{term\}} is the segmented target term (class), which is classified as positive terms (e.g., grass, garden, yard, etc.) and negative terms (e.g., building, person, car, etc.). Positive terms are SLZ target classes (e.g., grass, garden, water, patio, park, etc.) and the negative terms are unsafe landing zones (e.g. street, car, side walk, house, etc.). To simplify, let \emph{T} be the target classes and \emph{P} and \emph{N} be the positive and negative terms, respectively \eqref{eqn:terms}. In addition, \emph{a} and \emph{b} are the number of words in positive and negative terms, respectively \eqref{eqn:terms}. A single term is inserted into the pipeline \eqref{eqn:terms}.
\begin{figure*}[htbp]
\centerline{\includesvg[inkscapelatex=false, width=1\linewidth]{img/PEACE_Seg.svg}}
\caption{PEACE Segmentation Architecture: An observed aerial image and each words in target terms (positive and negative) are passed through the PEACE data pipeline to generate an optimal prompt, which is used to generate a segmentation per word. All segmentations generated using positive and negative terms are combined (union within each term) based on their corresponding terms (positive and negative). Final segmentation of the input image is generated by eliminating the combined segmentation of negative terms from the combined segmentation of the positive terms, removing the overlapping segmentations of negative terms. See equation \eqref{eqn:seg} for details.}
\label{methods:PEACE_Seg}
\end{figure*}
\begin{equation}
T =
\left(\begin{array}{cc}
P, N
\end{array}\right),
\quad
P = 
\left(\begin{array}{cc}
 p_1 \\
 p_2 \\
 \vdots \\
 p_{a-1} \\
 p_a \\
\end{array}\right),
\quad
N = 
\left(\begin{array}{cc}
 n_1 \\
 n_2 \\
 \vdots \\
 n_{b-1} \\
 n_{b} \\
\end{array}
\right)
\label{eqn:terms}
\end{equation} 

In summary, the PEACE data pipeline takes a target term and an observed aerial image to the modified CLIP Interrogator which has the image resolution, image type (frames), and environment description types. Then, the CLIP Interrogator generates best-matching words by ranking words based on word-to-image similarity (CLIP's cosine similarity) for each description types based on the image. The selected words are inserted into the optimal prompt (heuristically chosen the positioning of each description type through trial and error) with the target term to complete the prompt. This prompt and the input image are passed to an OV segmentation model to generate an SLZ segmentation heatmap.

\subsection{Combining Positive and Negative Term Segmentations}
For each word in terms \eqref{eqn:terms} for the input image, the PEACE data pipeline is used to generate a segmentation \emph{S} where \emph{Sp} and \emph{Sn} are segmentations of positive and negative terms, respectively, and \emph{Sf} is the final segmentation that combines all \emph{Sp} and eliminates \emph{Sn} \eqref{eqn:seg} (see Fig.~\ref{methods:PEACE_Seg} for visual).

\begin{equation}
    \begin{gathered}
        Sp = Sp_1 \cup Sp_2 \cup ... \cup Sp_{a-1} \cup Sp_a \\
        Sn = Sn_1 \cup Sn_2 \cup ... \cup Sn_{b-1} \cup Sn_b \\
        Sf = Sp \setminus Sn
    \end{gathered}
    \label{eqn:seg}
\end{equation}

\section{Experimental Setup}\label{expeimentalsetup}
\begin{figure*}[htbp]
\centerline{\includesvg[inkscapelatex=false, width=1\linewidth]{img/peace_path.svg}}
\caption{An example of a successful landing for PEACE-CS (DOVESEI-PEACE using CLIPSeg) Red circle is the initial starting point and the blue star is the final landing point. From right to left: initial UAV's view (altitude at 100m), zoom out with altitude 300m (trajectory in red, yellow dashed squre is the initial view), final location (altitude at 100m) and final UAV's view (altitude at 20m).}
\label{results:success}
\end{figure*}

To validate our PEACE system, we performed tests using DOVESEI~\cite{bong2023dynamic} as a baseline. We used high-resolution satellite images from Paris, France, sourced from our open-source ROS~2 package\footnote{\hyperlink{https://github.com/ricardodeazambuja/ROS2-SatelliteAerialViewSimulator}{https://github.com/ricardodeazambuja/ROS2-SatelliteAerialViewSimulator}} to simulate realistic conditions. Although we cannot guarantee absolute certainty, we believe that these specific images, paired with our prompts, were not encountered by the OV segmentation models during their training phase, thus constituting a zero-shot evaluation scenario. We use CLIPSeg (CS)~\cite{Luddecke_2022_CVPR}, a widely used OV segmentation model known for its relatively lower computational requirements, to evaluate whether implementing PEACE would improve SLZ segmentation performance. We compared our results with FastSAM (FS)~\cite{zhao2023fast} to demonstrate that PEACE can be applied to other OV models. We did not compare with traditional safe-landing methods since they are usually done at a lower altitude with senors, targeting a specific safe-landing area such as grass, flat surface or road, which is not an OV segmentation.
We conducted 100 identical safe landing tests on an Orin AGX for each configuration, with uniform starting conditions (altitude of 100m and consistent (x, y) positions). All configurations used DOVESEI's system framework for evaluation, but differed in prompt engineering approaches and segmentation methods (the latter only for PEACE). The three configurations compared were:
\begin{itemize}
\item DOVESEI-Default (DD), which uses CLIP's default prompt engineering (\emph{"A photo of {term}."}),
\item DOVESEI (D), which employs aerial-specific prompt engineering (\emph{"A bird's eye view of a {term}, in game screenshot, bad graphics, shade, shadows."}), and
\item DOVESEI-PEACE (PEACE), which implements our dynamic aerial prompt engineering (\emph{"A {resolution} {image type (frame)} of {term} in {environment}}.').
\end{itemize}

Therefore, we experimented with six methods using the three configurations for CLIPSeg (CS) and FastSAM (FS): DD-CS, D-CS, PEACE-CS, DD-FS, D-FS and PEACE-FS.
Following the methodology in~\cite{bong2023dynamic}, the simulated UAV always started at an altitude of 100m, its position being uniformly sampled within a predefined rectangular area formed by latitude and longitude coordinate pairs, and then descended to 20m. Landing goal positions were not predetermined, but rather visually identified during operation. Experiments exceeding the maximum allowed time (1200s) were automatically terminated.

\section{Results}
In order to validate the SLZ segmentation improvements of PEACE, we used the UAVid Semantic Segmentation Dataset~\cite{LYU2020108} to measure mIoU (mean Intersection over Union). The Paris satellite images were not used for mIoU because of the absence of ground truth segmentations. Using 400 aerial images, D-CS had 34 in mIoU, which is lower than DD-CS (38) and PEACE-CS had the highest mIOU (41) as observed in Table~\ref{results:aggregated_results}, which is a significant improvement as the improvement is purely based on prompt engineering. However, SLZ inference speed is lower for PEACE-CS compared to others, which is expected trade-off since the prompt is re-engineered for every image. Despite the latency in the SLZ inference speed, PEACE-CS' average time spent finding and descending to the SLZ is not as high, as illustrated in Table~\ref{results:aggregated_results}) since it is able to spot SLZs with higher confidence than other methods, which accelerates the overall time spent for final SLZ selection and descent.

\begin{table*}[htbp]
\centering
\captionsetup{justification=centering, belowskip=9pt}
\caption{Successful Safe-Landing for DD-CS vs D-CS vs PEACE-CS}
\renewcommand{\arraystretch}{1.5}
\begin{tabular}{|l|c|c|c|c|c|c|}
\hline
\multicolumn{1}{|c|}{} &
  \textit{\textbf{DD-CS}} &
  \textit{\textbf{D-CS}} &
  \textit{\textbf{PEACE-CS}} &
  \textit{\textbf{DD-FS}} &
  \textit{\textbf{D-FS}} &
  \textit{\textbf{PEACE-FS}}\\ \hline
\textit{\textbf{Total Successful SLZ selections}} & 76/100 & 57/100 & \underline{92/100} & 80/100 & 61/100 & \textbf{93/100} \\ \hline
\textit{\textbf{Average Horizontal Distance (m)}} & \underline{80.97}$\pm$20.11 & \textbf{74.40}$\pm$19.99 & 100.34$\pm$17.30 & 88.65$\pm$20.01 & 83.13$\pm$17.73 & 103.21$\pm$19.71 \\ \hline
\textit{\textbf{Average Time Spent (s)}} & \textbf{506.47}$\pm$49.30 & 843.98$\pm$66.98 & 593.68$\pm$52.75 & \underline{512.86}$\pm$62.91 & 901.04$\pm$70.12 & 611.92$\pm$55.31 \\ \hline
\textit{\textbf{mIoU}} & 38$\pm$2.33 & 34$\pm$2.10 & 41$\pm$1.42 &  \underline{43}$\pm$1.98 & 41$\pm$1.76 & \textbf{47}$\pm$2.45 \\ \hline
\textit{\textbf{Average Inference Speed (Hz)}} & 1.90$\pm$0.21 & 1.89$\pm$0.24 & 0.82$\pm$0.13 & \textbf{2.70}$\pm$0.26 & \underline{2.66}$\pm$0.30 & 0.78$\pm$0.11 \\ \hline
\end{tabular}
\subcaption*{Abbreviations: DD (DOVESEI-Default), D (DOVESEI), PEACE (DOVESEI-PEACE), CS (CLIPSeg), FS (FastSAM), m (meter), s (second), mIoU (mean Intersection over Union) and Hz (Hertz). Best results are \textbf{bolded} and second best results are \underline{underlined}.}
\label{results:aggregated_results}
\end{table*}

\begin{figure}[htbp]
\centerline{\includegraphics[width=1\linewidth]{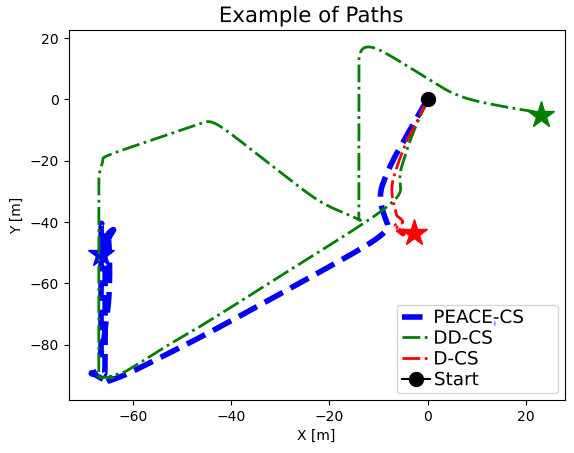}}
\caption{Paths generated for landing experiments starting at the same location, where stars indicate safe-landing zone selection reached at the end of exploration. Abbreviations: DD (DOVESEI-Default), D (DOVESEI), PEACE (DOVESEI-PEACE) and CS (CLIPSeg).}
\label{results:paths}
\end{figure}

\begin{figure}[htbp]
\centerline{\includegraphics[width=1\linewidth]{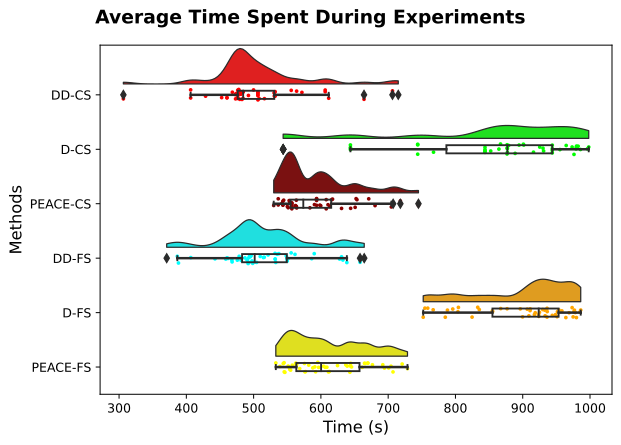}}
\caption{Average Time spent from initial starting point to descent until 20m after finding a safe-landing zone for each experiment. Abbreviations: DD (DOVESEI-Default), D (DOVESEI), PEACE (DOVESEI-PEACE), CS (CLIPSeg) and FS (FastSAM).}
\label{results:time}
\end{figure}

From 100 identical safe-landing experiments using Paris satellite images, PEACE-CS had 92/100 successful safe-landing zone detection (Fig.~\ref{results:success} show a successful landing example), performing significantly better than D-CS (57/100) and DD-CS (76/100) (Table~\ref{results:aggregated_results}). The reasoning behind the higher average time spent by D-CS is its prompt engineering, which made the segmentation more brittle and leading the system to be less confident. More specifically, optimized prompt engineering can make the segmentation more stable and increase the system's confidence, therefore it would not have to go to the waiting state to figure out a better SLZ. The longer average horizontal distance for PEACE-CS is due to finding a safer SLZ that is further away (Fig.~\ref{results:paths}) or finding a better SLZ during descent and adjusting its path, which is out of D-CS and DD-CS SLZ selections capabilities since their prompts are static. We also realized that PEACE-CS found SLZs that are closer to its location that other methods could not. This reasoning can also explain the varying time spent on experiments (Fig.~\ref{results:time}) for PEACE and its total successful SLZ selections. It is also important to mention that shorter time spent during exploration, shorter horizontal distance, and inference speed does not always indicate better results since some experiments chose incorrect SLZs, but faster. Thus, the most important metric for safe-landing is the total successful SLZ selections, for which PEACE shows the best results.
\begin{figure}[htbp]
\centerline{\includegraphics[width=1\linewidth]{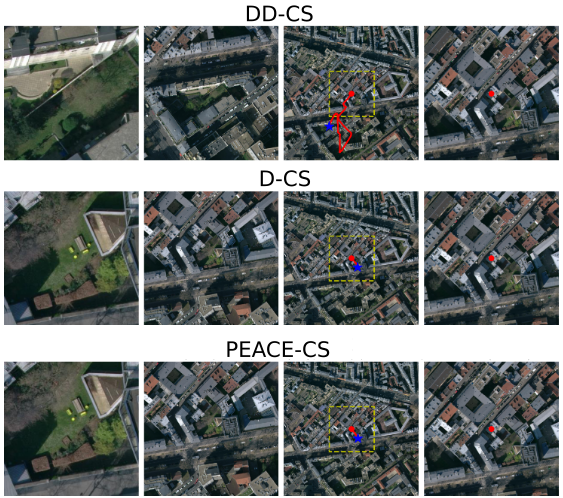}}
\captionsetup{belowskip=0pt}
\caption{Paths generated for landing experiments starting at the same location, where stars indicate where each trajectory reached at the end of the experiment. Abbreviations: DD (DOVESEI-Default), D (DOVESEI), PEACE (DOVESEI-PEACE) and CS (CLIPSeg)}
\label{results:paths_diff}
\end{figure}
Compared to DD-CS, D-CS and PEACE-CS selected similar SLZs (Fig.~\ref{results:paths_diff}). This is due to both having aerial prompt engineering, compared to the generic prompt engineering of DD-CS. Despite choosing similar SLZs, PEACE-CS had much higher success rate (Table~\ref{results:aggregated_results}) than DOVESEI due to its precision on segmentation during descent, which is only possible due to the automated dynamic prompt engineering that is reactive to changing environments (images). This precision adjustment from PEACE justifies longer average horizontal distance in Table~\ref{results:aggregated_results}.

\begin{figure}[htbp]
\centerline{\includesvg[inkscapelatex=false, width=1\linewidth]{img/FastSAMDiff.svg}}
\caption{Differences in segmentation of grass using prompt engineering from DD-FS, D-FS and PEACE-FS. Only PEACE-FS differs in prompts due to its dynamic prompt engineering feature. Generated prompts from PEACE-FS in the order of images starting from the top to bottom: \emph{"A \{rendered\} \{image\} of \{grass\} in \{spring\}}.", \emph{"A \{3D\} \{art\} of \{grass\} in \{spring\}}.", \emph{"A \{rendered\} \{view\} of \{grass\} in \{autumn\}}." and \emph{"A \{blurred\} \{view\} of \{grass\} in \{map\}}.". Abbreviations: DD (DOVESEI-Default), D (DOVESEI), PEACE (DOVESEI-PEACE) and FS (FastSAM)}
\label{results:fastSAMDiff}
\end{figure}
We also conducted the safe-landing experiments using FastSAM (FS)~\cite{zhao2023fast} instead of CLIPSeg (CS) as a baseline to demonstrate the adaptability of PEACE to other common OV models. FS was chosen because it is a widely used model that is not only known for its accuracy but also fast, which is essential for an onboard system mounted on a UAV. This experiment was done in the exact same setting using the same images as the previous experiments done using CS for a fair comparison. Figure~\ref{results:fastSAMDiff}, shows the visual differences in segmentation generated using different prompt engineering methods and how PEACE-FS segments better than DD-FS and D-FS.

\begin{figure}[htbp]
\centerline{\includegraphics[width=1\linewidth]{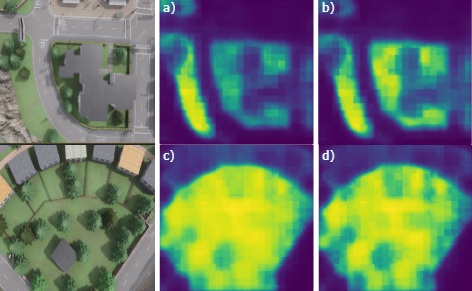}}
\caption{An example of segmentation difference between DD-CS and PEACE-CS: 
a) \emph{"A photo of \{grass\}}.";
b) \emph{"A \{blurry\} \{photo\} of \{grass\} in \{autumn\}}.";
c) \emph{"A photo of \{grass\}}.";
d) \emph{"A \{3D\} \{photo\} of \{grass\} in \{morning\}}."}
\label{results:diff}
\end{figure}

The visual differences were validated by generating the mIoU of SLZ segmentations using the same UAVid Semantic Segmentation Dataset used for PEACE-CS. Table~\ref{results:aggregated_results} demonstrates that PEACE-FS had an mIoU of 47, confirming that the dynamic prompt engineering of PEACE is advantageous, compared to DD-FS and D-FS, which had 43 and 41, respectively. Despite CS being a less precise segmentation method compared to FS, it is worth mentioning that PEACE-CS was able to achieve the same mIoU of 43 as D-FS without any learning or optimization (just dynamic prompt engineering). Generally, FS improved the successful safe-landing due to its higher mIoU compared to CS, PEACE-FS shows some diminishing returns: PEACE-FS also had 93/100 successful safe-landing (Table~\ref{results:aggregated_results}), similar average horizontal distance and time spent for finding a SLZ and landing. It is important to note that PEACE's dynamic prompt engineering still produced better SLZ selection and landing compared to other methods for both CS and FS based experiments. 
Before proceeding to real-world image testing, we validated our hypothesis in a virtual environment (CARLA~\cite{carla_web}) and observed significant segmentation improvements using the PEACE prompt engineering approach compared to other methods, as illustrated in Fig.~\ref{results:diff}.
\subsection{Limitations}
\subsubsection{Limits of Prompt Engineering}
We believe that prompt engineering can improve aerial segmentation to a certain point, but additional optimization and modifications to the models (e.g., fine-tuning, learning, etc.) and image processing methods (e.g., reusing features, etc.) are necessary to improve SLZ segmentation beyond the default capabilities of OV models.

\subsubsection{Real Robot Experiments}
\begin{figure}[htbp]
\centerline{\includegraphics[width=.7\linewidth]{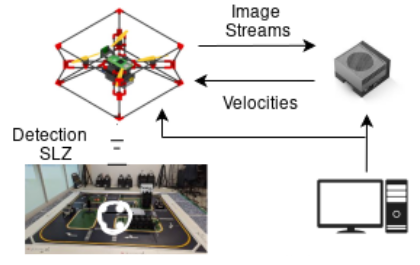}}
\caption{System Architecture for the indoor experiments: Personal computer was used to start our experiments. Orin AGX was used to do all computation, but not mounted to the UAV due to its weight (a lightweight indoor UAV was used for safety). A toy town was used as a map.}
\label{figure:indoor}
\end{figure}

Due to the legal and safety restrictions, we were not able to conduct outdoor experiments and run our experiments in an indoor setting (Fig.~\ref{figure:indoor}), which can be very limited in terms of imitating a real-world environment. Ten safe-landing experiments were conducted on a town map, using a Cognifly~\cite{de_Azambuja_2022} UAV. Optitrack~\cite{optitrack} was used for positioning, and Orin AGX performed all PEACE calculations. PEACE-CS had a successful selection of SLZ of 10/10, compared to DD-CS and D-CS, which had 7 and 6, respectively. 
It is important to mention that Orin AGX was not mounted on the UAV due to the Cognifly's small payload capacity, which is reasonable for indoor experiments. In an outdoor setting, the Orin AGX would be mounted directly on a UAV. In addition, the map used for this experiment is simpler than an urban city environment at 100m altitude and an outdoor validation is an important milestone yet to be achieved. Given the nature of volatility in visual-language models and the safety of landing operations, rigorous outdoor experiments in complex environments are critical to justify the real-world usage. We also did not test on a simulator due to its lack of realistic features, but it would be advantageous to test the environmental diversity (weather, lighting, etc.) near future. 

\section{Conclusions}
In this paper, we have introduced PEACE (Prompt Engineering Automation for CLIPSeg Enhancement), a novel approach for aerial image segmentation that significantly advances our previous work on Dynamic Open-Vocabulary Enhanced Safe-landing with Intelligence (DOVESEI). PEACE addresses key limitations in open-vocabulary segmentation for UAV applications through two major innovations: automated dynamic prompt generation tailored to each input frame and an intelligent segmentation strategy that combines positive and negative term classifications to refine landing zone identification.
Our experimental evaluation demonstrates the substantial performance improvements achieved by PEACE. When integrated with CLIPSeg (CS) and applied to the DOVESEI framework, PEACE-CS dramatically increased successful landing attempts from 57/100 to 92/100, representing a 62\% improvement in reliability. This marked enhancement in performance underscores the critical importance of context-aware, adaptive prompt engineering for real-world aerial robotics applications, particularly in safety-critical scenarios like emergency landings.

The results suggest that PEACE not only improves segmentation accuracy but also enhances system robustness across varying conditions. As autonomous aerial systems continue to evolve for deployment in complex urban environments, approaches like PEACE that can dynamically adapt to changing visual contexts without requiring extensive retraining will be increasingly valuable. Future work will focus on further optimizing computational efficiency for resource-constrained UAV platforms and expanding the system's capabilities to handle more diverse and challenging landing scenarios.








\bibliography{iros2025}
\bibliographystyle{IEEEtran}

\end{document}